\documentclass[runningheads]{llncs}

\usepackage{eccv}

\usepackage{color, colortbl}
\usepackage{pifont}
\usepackage{multirow}
\usepackage[symbol]{footmisc}

\newcommand{\parsection}[1]{\noindent\textbf{#1}:}
\definecolor{Gray}{gray}{0.8}
\definecolor{darkgreen}{RGB}{0,150,0} % You can adjust the RGB values

\usepackage{eccvabbrv}

\usepackage{graphicx}
\usepackage{booktabs}
\usepackage[accsupp]{axessibility}  % Improves PDF readability for those with disabilities.

\usepackage[pagebackref,breaklinks,colorlinks,citecolor=eccvblue]{hyperref}
\usepackage{orcidlink}

\begin{document}

\title{NeuroNCAP: Photorealistic Closed-loop Safety Testing for Autonomous Driving} 

\titlerunning{NeuroNCAP}

% TODO: Use this for orcid icon
% \author{William Ljungbergh\inst{*,1,2}\orcidlink{0000-0002-0194-6346} \and
% Adam Tonderski\inst{*,1,3}\orcidlink{0000-0002-2160-4386} \and
% Joakim Johnander\inst{1}\orcidlink{0000-0003-2553-3367} \and
% Holger Caesar\inst{4}\orcidlink{0000-0001-5099-6297} \and
% Kalle Åström\inst{3}\orcidlink{0000-0002-8689-7810} \and
% Michael Felsberg\inst{2}\orcidlink{0000-0002-6096-3648} \and
% Christoffer Petersson\inst{1}\orcidlink{0000-0002-9203-558X}
% }

% TODO: Use for non-orcid icon
\author{William Ljungbergh\inst{*,1,2} \and
Adam Tonderski\inst{*,1,3} \and
Joakim Johnander\inst{1} \and
Holger Caesar\inst{4} \and
Kalle Åström\inst{3} \and
Michael Felsberg\inst{2} \and
Christoffer Petersson\inst{1}
}
\authorrunning{W. Ljungbergh et al.}

\institute{
    Zenseact, Gothenburg, Sweden \and
    Linköping Univeristy, Linköping, Sweden \and
    Lund University, Lund, Sweden \and
    Delft University of Technology, Delft, Netherlands
}
\maketitle
% Added to show equal contribution
\footnotetext[1]{Denotes equal contribution.}
\begin{center}
\href{https://github.com/atonderski/neuro-ncap}{\texttt{https://github.com/atonderski/neuro-ncap}
}
\end{center}
\begin{abstract}
We present a versatile NeRF-based simulator for testing autonomous driving (AD) software systems, designed with a focus on sensor-realistic closed-loop evaluation and the creation of safety-critical scenarios. The simulator learns from sequences of real-world driving sensor data and enables reconfigurations and renderings of new, unseen scenarios. In this work, we use our simulator to test the responses of AD models to safety-critical scenarios inspired by the European New Car Assessment Programme (Euro NCAP). Our evaluation reveals that, while state-of-the-art end-to-end planners excel in nominal driving scenarios in an open-loop setting, they exhibit critical flaws when navigating our safety-critical scenarios in a closed-loop setting. This highlights the need for advancements in the safety and real-world usability of end-to-end planners. By publicly releasing our simulator and scenarios as an easy-to-run \href{https://github.com/atonderski/neuro-ncap}{evaluation suite}, we invite the research community to explore, refine, and validate their AD models in controlled, yet highly configurable and challenging sensor-realistic environments. 
%End-to-end planning -- where a neural network is trained to map sensor input to a trajectory to follow -- has recently achieved outstanding results on standard Autonomous Driving dataset benchmarks. However, these benchmarks fail to take two aspects into account: the closed-loop nature of the problem; and the very rarely occurring safety-critical scenarios.  In this work, we propose a NeRF-based simulator in which automotive driving systems can be evaluated in a closed-loop setting. The simulator enables us to create safety-critical scenarios, inspired by the industry standard European New Car Assessment Programme (Euro NCAP), by inserting objects that will cause a collision unless avoided. We find that state-of-the-art end-to-end planners fail in safety-critical scenarios, e.g., driving straight into a stationary vehicle placed in the ego lane. By publicly releasing our simulator and safety-critical scenarios as an easy-to-run evaluation suite, we hope to accelerate the refinement and validation of end-to-end planners.
\end{abstract}

\begin{figure}[t]
    \centering
    \includegraphics[width=\textwidth]{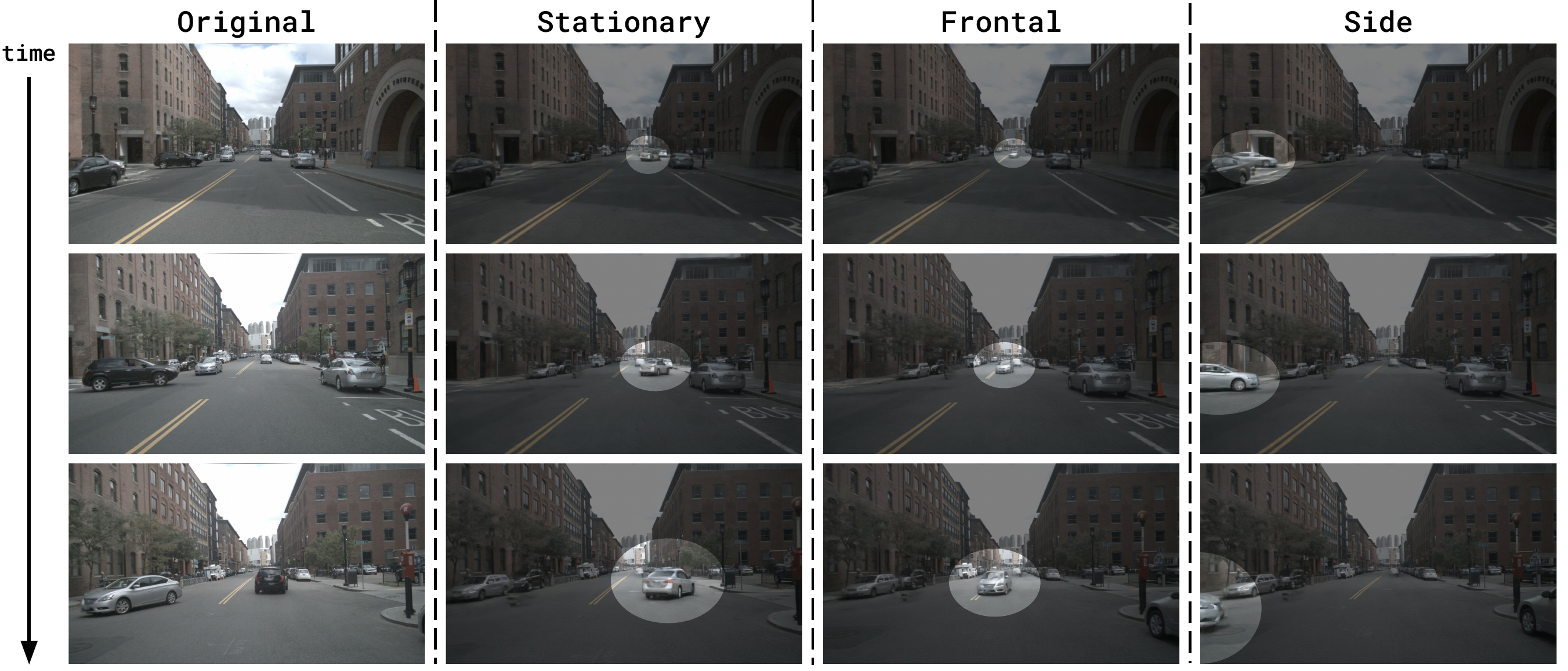}
    \caption{The core idea in NeuroNCAP is to leverage NeRFs to realistically simulate many safety-critical scenarios from a sequence of real-world data. Here we show the original scenario, followed by examples of our three types of collision scenarios: stationary, frontal, and side. The inserted safety-critical actor has been highlighted for illustration purposes. We can generate hundreds of unique scenarios from each log by selecting different actors, jittering their trajectories, and choosing different starting conditions for the ego vehicle. Note that scenarios are not pre-generated, but rather obtained by iteratively generating new images, computing a plan, and acting upon said plan.}
    \label{fig:frontpage}
\end{figure}

\section{Introduction}
\label{sec:intro}
\phantom{} 

Recent work on autonomous driving (AD)~\cite{hu2023planning,jiang2023vad} suggests designing and training a holistic neural network for mapping sensor inputs directly to a planned trajectory. Compared to prior work that used modular software stacks, engineered interfaces between modules, or handcrafted rules, this end-to-end approach has several advantages. First, as the driving behavior is learned, the predicted trajectories are expected to resemble how a typical human driver would act. Second, the approach is scalable in the sense that more data leads to more robust as well as generalizable driving performance~\cite{bojarski2016end,jiang2023vad} and in the sense that there is no need to manually design intermediate interfaces or cost functions. The neural network may be divided into modules, but the interfaces between them are learned in order to mitigate information loss.
%~\cite{bhat2020learning}
%While the neural network can be divided into modules, the interfaces between them use representations in some learned, latent space. This allows the neural network to learn what information is necessary to propagate and thus avoid information loss.
%Moreover, while the neural network can be divided into modules, these modules are trained end-to-end to avoid error propagation. The interfaces between these modules use representations in some learnt, latent space, which allows the neural network to learn what information is necessary to propagate and could thus avoid information loss.

Hu \etal~\cite{hu2023planning} demonstrated that their end-to-end planner, UniAD, performed well on the popular nuScenes~\cite{caesar2020nuscenes} planning benchmark. This is an open-loop benchmark, where the tested planner never influences the driving. Instead, the plans are compared to the trajectory taken by the vehicle during data collection and a score is computed based on the similarity between the two. Codevilla \etal~\cite{codevilla2018offline}, as well as Dauner \etal~\cite{dauner2023parting}, shed some doubt about the correlation between such an open-loop score and the actual driving performance. This begs the question, how would state-of-the-art end-to-end planners fare if their predicted policy would be acted upon? Unlike regular planners that can be evaluated in a closed-loop manner using straightforward object-level simulations, end-to-end planners require complex sensor simulations to accurately predict their behavior in real-world scenarios. This introduces significant challenges due to the complexity and computational demands of high-fidelity sensor simulation. Moreover, the nuScenes benchmark contains normal driving scenarios, in which no collisions occur. It is unclear how state-of-the-art end-to-end planners would perform in safety-critical scenarios, where a crash is likely unless swift corrective action is undertaken.

In this work, we subject state-of-the-art end-to-end planners to closed-loop evaluation in safety-critical scenarios. Given sensor data, planners predict a plan. The plan is then executed under the constraints of a vehicle model in order to propagate the state of the ego-vehicle forward in time. Given the new state, we use recent advances in neural rendering -- NeRFs -- to resolve the problem of generating realistic sensor data. These three steps are then repeated until either a crash occurs or we deem the scenario to be over. By executing the predicted plan, we aim to reduce the gap between model evaluation and deployment. It is possible that the neural renderer exacerbates the gap, if the rendered images are not sufficiently realistic. We quantify to what extent the renderer affects the gap by analyzing the intermediary perception outputs that are available in state-of-the-art (SotA) end-to-end planners, such as UniAD~\cite{hu2023planning} and VAD~\cite{jiang2023vad}. Moreover, we ensure that replacing the real sensor data with rendered sensor data during open-loop evaluation does not negatively affect the open-loop performance.

To generate safety-critical scenarios, we take inspiration from the European New Car Assessment Protocol (Euro NCAP) for collision avoidance~\cite{euroNCAP2023collision}. This protocol comprises several scenario types that have been identified as safety-critical. These scenario types are rare, but are likely to lead to a collision unless the planner properly deals with them. We craft scenarios by altering recordings of scenes from the nuScenes dataset~\cite{caesar2020nuscenes}. We evaluate the driving quality by whether there is a crash, and at what velocity that crash occurs. Our benchmark should be viewed as a necessary but not sufficient condition for high quality driving. To summarize, our contributions are as follows:
\begin{enumerate}
    \item We release an open source framework for photorealistic closed-loop simulation for autonomous driving.
    \item We construct safety-critical scenarios, inspired by the industry standard Euro NCAP, that cannot safely be collected in the real world.
    \item Using the simulator and our scenarios, we design a novel evaluation protocol that focuses on collisions rather than displacement metrics.
    \item We show that two SotA end-to-end planners fail severely in our safety-critical scenarios despite accurately perceiving the environment.
    %\item We show that the perception of tested SotA end-to-end planners can correctly gauge the surrounding environment when presented with the rendered input, but are unable to deal with the domain gap between nominal and safety-critical driving.
    %\item When testing two SotA models for end-to-end autonomous driving using our protocol, we find that the perception can correctly gauge the surrounding environment when presented with the rendered input, but the planner is unable to deal with the domain gap between nominal and safety-critical driving.
\end{enumerate}

%We remove dynamic objects and select one at random to constitute a target actor. This target actor is then placed and moved according to a Euro NCAP scenario type. Planners are then subject to the altered scene and closed-loop evaluation. Following Euro NCAP, we evaluate the driving quality by whether there is a crash, and at what velocity the crash occurs. Our benchmark should be viewed as a necessary but not sufficient condition for safe driving quality.

%figure showing WHAT we do, not how (TODO: how this figure looks)
\section{Related Work}
\label{sec:relatedwork}

% uniad, vad, etc. modular vs end-to-end. benefits, drawbacks.
\parsection{End-to-end driving models} 
The autonomous driving task has traditionally been divided into different modules -- e.g., perception, prediction, and planning -- that are constructed individually~\cite{hu2023planning,jiang2023vad,mobileye2024ces}. Hu \etal~\cite{hu2023planning} argue that this division comes with a number of disadvantages: information loss across modules, error accumulation, and feature misalignment. Jiang \etal~\cite{jiang2023vad} emphasize that a planning module might need access to the semantic information of the sensor data that would not be present in a hand-crafted interface. These two works proceed to argue in favor of end-to-end planning. The pioneering work of Pomerlau \etal~\cite{pomerleau1988alvinn} proposes such a planner, where a single neural network is trained to map sensor input to an output trajectory. Decades of neural network advancements sparked new interest in end-to-end planning~\cite{chen2015deepdriving,bojarski2016end,codevilla2018end,codevilla2019exploring,prakash2021multi}. The black-box nature of these planners, however, makes them difficult to optimize and their results hard to interpret~\cite{jiang2023vad}. Hu \etal~\cite{hu2023planning} and Jiang \etal~\cite{jiang2023vad} propose two end-to-end neural network planners with intermediary outputs, corresponding to those of a modular approach. Their planners are divided into modules, but the module interfaces are learned, consisting of deep feature vectors.

\parsection{Open-loop evaluation of end-to-end planners}
Pomerleau \etal~\cite{pomerleau1988alvinn} evaluated their driving model by letting it drive a real-world test-vehicle. Such a setup makes large-scale testing costly and results may be hard to reproduce. Recent works on end-to-end planning~\cite{hu2023planning,jiang2023vad} instead evaluate in \emph{open loop}, where models predict a plan based on recorded sensor data. The predicted plans are never executed and instead, the actions are fixed to whatever was recorded. This setup has also been used in works on object-level planning~\cite{gao2020vectornet,liang2020learning,deo2022multimodal}, which assume perfect perception and feed both a map of the static environment as well as tracks of dynamic objects into the model. Such open-loop evaluation constitutes a gap between evaluation and real-world deployment. Moreover, performance is usually measured as the distance between the predicted plan and the trajectory driven by the vehicle in the recording~\cite{hu2023planning,jiang2023vad,gao2020vectornet,liang2020learning,deo2022multimodal}. While an error of zero corresponds to human-level driving, it is not necessarily true that a lower error is better. This can be realized by considering a scenario where two different trajectories are equally good. These issues were studied by Codevilla \etal~\cite{codevilla2018offline} and they found that open-loop evaluation is not necessarily correlated with actual driving quality. Dauner \etal~\cite{dauner2023parting} draw similar conclusions.

% Object level closed loop, CARLA (and proprietary), VISTA, nuPlan~\cite{caesar2021nuplan,karnchanachari2024nuplan}, NeRF, NeuRAD, etc.

% Holger: Different types of simulators:
%- Box-based simulators: nuPlan, Waymax, TUM Common Road, Sumo etc. --> don't generate sensor data
%- Hand-crafted graphical simulators: Carla, AirSim (mostly drones), Siemens Prescan etc. --> generate sensor data, but not photorealistic; require extensive modelling efforts to represent the environment.
%- Dreamed simulators: Dreamer v1/v2/v3 etc. --> World model can predict future in a latent space --> no need for photo-realism, but cannot directly be used for end-to-end planners, unless we generate images. (which leads to the NeRF-style simulators below) Contrary to all other approaches, the scene is not editable in the latent space, at least not in an intuitive way.
%- Projection-based simulators: VISTA etc., --> use simple transformations to synthesize views from novel-viewpoints, only works for small transformations.
%- NeRF-based simulators: NeuRAD, UniSim etc., --> allow for vastly larger "baselines" when synthesizing novel viewpoints. Photo-realistic. 
%Somebody should check where Wayve's Gaia-1 world model + image generation ends up. 
% Holger: It is difficult to disentangle evaluation from simulation. E.g. closed-loop simulation has different metrics from open-loop simulation. Perhaps the previous section could only deal with evaluation in open-loop and this section deals with evaluation in closed-loop, which necessitates simulation?
\parsection{Closed-loop evaluation and simulation} 
Given the aforementioned issues of open-loop evaluation, closed-loop simulation becomes attractive. Several object-level simulators have been proposed~\cite{krajzewicz2010traffic,althoff2017commonroad,caesar2021nuplan,gulino2024waymax}. These simulators do not generate sensor data, however, which makes it impossible to test end-to-end planners in closed loop. A number of hand-crafted graphical simulators have been proposed~\cite{dosovitskiy2017carla,shah2018airsim,son2019simulation}. The challenge for such simulators is twofold: it is difficult to create realistic-looking images and it is hard to create graphical assets that capture the variety of the real world. Work on world models~\cite{watter2015embed,hafner2019dream} demonstrate that the future of a scene -- e.g., an Atari game -- can be predicted in latent space and that vectors in latent space can be decoded into sensor input. Hu \etal~\cite{hu2023gaia} build a world-model from a large-scale real-world automotive dataset. Amini \etal~\cite{amini2020learning} propose VISTA, in which novel views can be synthesized around the local trajectory by unprojecting the closest image via predicted depth, and reprojecting. Yang \etal~\cite{yang2023unisim} propose to use neural radiance fields (NeRF) to create photorealistic sensor input of a scene. The method was subsequently improved by Tonderski \etal~\cite{tonderski2023neurad}  with more accurate sensor modeling and higher rendering quality, particularly for the 360-degree setting considered here.

%\parsection{Euro NCAP}
%The European New Car Assessment Programme (Euro NCAP) constitute an industry standard for vehicle safety~\cite{van2016european}. This program covers aspects beyond automated driving, such as pedestrian protection or speed assistance systems, but in this work we primarily make use of the Euro NCAP automatic collision avoidance assessment protocol~\cite{euroNCAP2023collision}. This protocol provides scenarios in which a crash will occur unless action is taken. To obtain a full score, the vehicle needs to brake or steer to avoid the accident. Partial scores can be obtained if the impact velocity is sufficiently reduced.

\parsection{New Car Assessment Programs}
The New Car Assessment Program (NCAP) was introduced in 1979 by the U.S. Department of Transportation's National Highway Safety Administration in order to provide consumers with information on the relative safety potential of automobiles~\cite{hershman2001us}. NCAP crash-tested vehicles and scored vehicles based on the probability of serious injury. A similar European protocol was proposed in 1996, the European New Car Assessment Programme (Euro NCAP). In 2009, Euro NCAP was overhauled in order to also include tests of emerging crash avoidance systems~\cite{van2016european}. Initially, this included electronic stability control and speed assistance systems, but this was later extended to include additional systems, such as autonomous emergency braking~\cite{van2016european} and autonomous emergency steering~\cite{euroNCAP2023collision}. In this work, we take inspiration from the Euro NCAP automatic collision avoidance assessment protocol~\cite{euroNCAP2023collision}. This protocol provides scenarios in which a crash will occur unless action is taken. To obtain a full score, the vehicle needs to brake or steer to avoid the accident. Partial scores are awarded if the impact velocity is sufficiently reduced.

\section{Method}  % Safety-Critical Evaluation Protocol?
Our end-to-end planning evaluation protocol comprises a closed-loop-simulator (see Section~\ref{sec:closed-loop-sim}) and a collision-focused evaluation protocol (see Section~\ref{sec:evaluation}).

%figure showing how neurad interacts with models and nerfs (include vehicle model, collision check).

\subsection{Closed-loop Simulator}  % describe figure 2
\label{sec:closed-loop-sim}
\begin{figure}[t]
    \centering
    \includegraphics[width=0.9\textwidth]{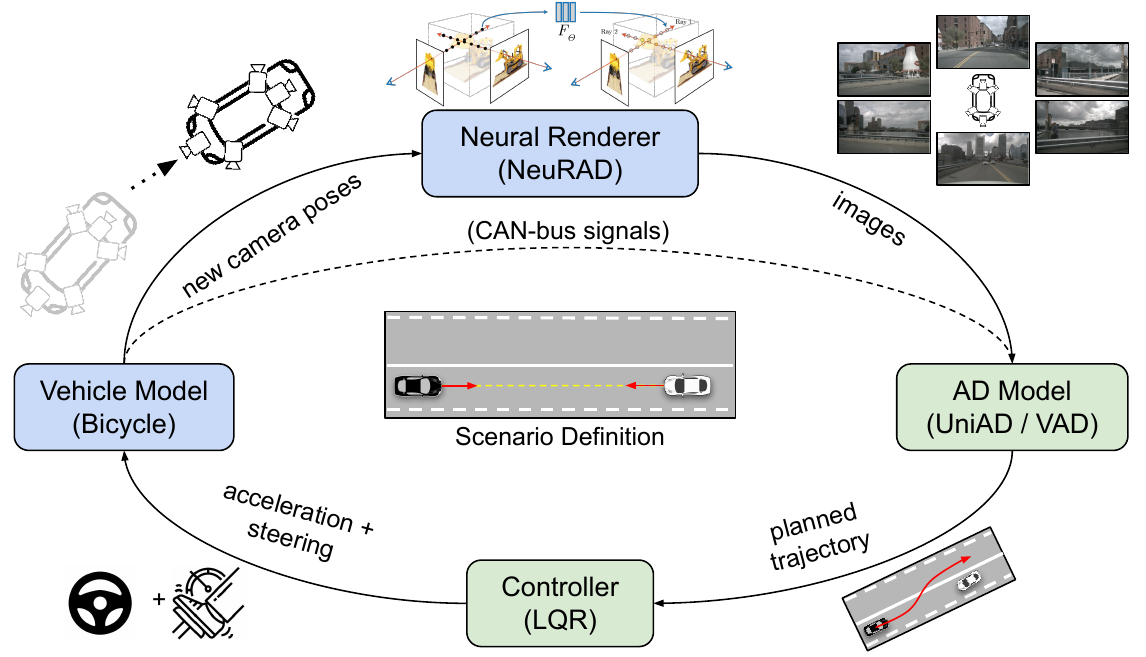}
%    \caption{Overview of our closed loop simulation engine (PLACEHOLDER).}
    \caption{Our closed-loop simulation engine comprises four parts. First, given a driving log, a neural renderer (NeRF) provides photo-realistic images given the ego-vehicle state. Second, an AD model (e.g., the end-to-end planner UniAD~\cite{hu2023planning}) uses these to predict a future ego-trajectory. Third, a controller estimates acceleration and steering signals. Finally, a vehicle model propagates the ego-vehicle state one step into the future. This process is then iterated to achieve closed-loop simulation. Blue indicates simulator, green indicates AD system.}
    \label{fig:overview}
\end{figure}

Our closed-loop simulator repeatedly performs four steps. First, high-quality camera input is rendered given the state and camera calibration of the ego vehicle. The renderer is constructed from a log of a driving vehicle. Second, the end-to-end planner predicts a future ego-vehicle trajectory given the rendered camera input and the ego-vehicle state. Third, a controller converts the planned trajectory to a set of control inputs. Fourth, a vehicle model propagates the ego-state forward in time given the control inputs. This procedure is illustrated in Figure~\ref{fig:overview}. Next, we elaborate on each of the four steps.

\parsection{Neural renderer}
In order to simulate novel sensor data, we adopt a neural renderer~\cite{mildenhall2021nerf}. NeRFs learn an implicit representation of the 3D environment from a log of collected real-world data. Once trained, NeRFs can render sensor-realistic novel views from said scene. Recent advances add the ability to also edit the dynamic objects in the scene by changing their corresponding 3D bounding boxes~\cite{ost2021neural}. Specifically, actors can be removed, added, or set to follow novel trajectories, which in our case enables the creation of safety-critical scenarios. For example, to simulate a rare but critical safety scenario, a vehicle that is originally moving in an adjacent lane can be positioned to be stationary and in the same lane as the ego-vehicle. This novel situation necessitates the ego-vehicle to either brake or execute a precise overtaking maneuver.

Two things should be noted. First, the recently proposed NeuRAD~\cite{tonderski2023neurad} also supports the rendering of LiDAR data. However, as state-of-the-art end-to-end planners consume only camera data, we focus only on camera data in this work. Second, as we show in our experiments, the domain gap introduced by modern NeRFs compared to real data is sufficiently small for the perception parts of end-to-end planners to still function with high performance. However, we expect this gap to be reduced further with future developments in neural rendering.

%\red{While NeuRAD supports rendering both images and LiDAR data, we have in this work only focused on methods that only use images as input}

\parsection{AD model}
Recent works on end-to-end planning~\cite{hu2023planning, jiang2023vad, hu2022st} describe a system that consumes (i) raw sensor data; (ii) the ego-vehicle state; and (iii) a high-level plan to predict a planned trajectory. The planned trajectory comprises waypoints at some frequency and with some time horizon.  It should be noted that while we primarily aim to analyze state-of-the-art end-to-end planners, this module can be replaced with any type of planner, e.g., a modular detector-tracker-planner pipeline.

\parsection{Controller}
In order to apply the vehicle model, the waypoints need to be converted to a sequence of control signals, corresponding to a sequence of steering angle ($\delta$) and acceleration ($a$) commands. Following Caesar \etal~\cite{karnchanachari2024nuplan}, we achieve this with a linear quadratic regulator (LQR). Note that while we only analyze planners that output waypoints, the planner could instead directly output a sequence of control signals.
% An end-to-end planner incorporates a system which consumes raw sensor data and a high-level plan as input and outputs either a planned trajectory (i.e., timed waypoints) or a set of control inputs (e.g., steering rate and acceleration). Note that if the system output timed future waypoints -- as is the case for many SOTA end-to-end driving models -- they have to be converted to control signals to enable the  propagation of our current state (via the vehicle modeling described below). For fairness and simplicity we rely on previous work~\cite{caesar2021nuplan} and use a Linear Quadratic Regulator (LQR) to compute control inputs from a sequence of waypoints.
% Note that our framework makes no assumptions on the planning system -- it can be a modular detection-tracking-planning model as well as an end-to-end driving model. 

\parsection{Vehicle model}
Given a set of control signals, generated from the planned trajectory, the vehicle state is propagated through time. To this end, we follow prior closed-loop simulators~\cite{caesar2021nuplan,karnchanachari2024nuplan} and adopt a discrete version of the kinematic bicycle model~\cite{rajamani2011vehicle}. It can formally be described as

\begin{equation}
S = \begin{pmatrix} x \\ y \\ \theta \\ v \end{pmatrix}, \quad
\frac{dS}{dt} = 
\begin{pmatrix}
    v\cos{\theta} \\
    v\sin{\theta} \\
    \frac{v\tan{\delta}}{L} \\
    a
\end{pmatrix}\enspace.
\end{equation}
The state $S$ is composed of $x$, $y$, $\theta$, and $v$, where $x$ and $y$ is the longitudinal and lateral position; $\theta$ the rotation; and $v$ the speed of the vehicle. Furthermore, $L$ is the wheelbase of the vehicle, and $\delta$ and $a$ are the control signals. We adopt control signal limits and the wheelbase based on the vehicle that was used when collecting the data. Note that $x$, $y$, and $\theta$ live in a global coordinate system, whereas $v$, $L$, $\delta$, and $a$ are values.

\subsection{Evaluation}
\label{sec:evaluation}
In contrast to common evaluation practices -- \ie, averaging performance across large-scale datasets -- we instead focus our evaluation on a small set of carefully designed safety-critical scenarios. These scenarios have been crafted such that any model that cannot successfully handle all of them, should be considered unsafe. We have taken inspiration from the industry standard Euro NCAP testing~\cite{euroNCAP2023collision} (see Section~\ref{sec:relatedwork}) and define three types of scenarios, each characterized by the behavior of the actor that we are about to collide with: \emph{stationary}, \emph{frontal}, and \emph{side}. Following the Euro NCAP nomenclature, we refer to this actor as the \emph{target actor}. The aim is to control the ego-vehicle to avoid a collision with the target actor or at least decrease the collision velocity.

For each scenario type, we create multiple scenarios. Each scenario is based on data collected from around 20 seconds of real-world driving. The ego-vehicle and target actor states are initialized such that if current speeds and steering angles are maintained, a collision will occur at approximately 4 seconds into the future. All non-stationary actors are removed from the scene and we randomly select one of these to be the target actor, taking into consideration whether the actor has been observed sufficiently closely, and under the necessary angles, to produce realistic renderings. As our renderer is limited to rigid actors, we exclude pedestrians from this selection. Finally, we randomly jitter the position, rotation, and velocity of the target actor within scenario-specific intervals. During evaluation, we run each scenario for a large number of runs (with a fixed random seed) and compute average results. Next, we describe the characteristics of each type of scenario.

\begin{figure}[t]
\centering
\begin{subfigure}{0.32\textwidth}
    \includegraphics[width=\textwidth]{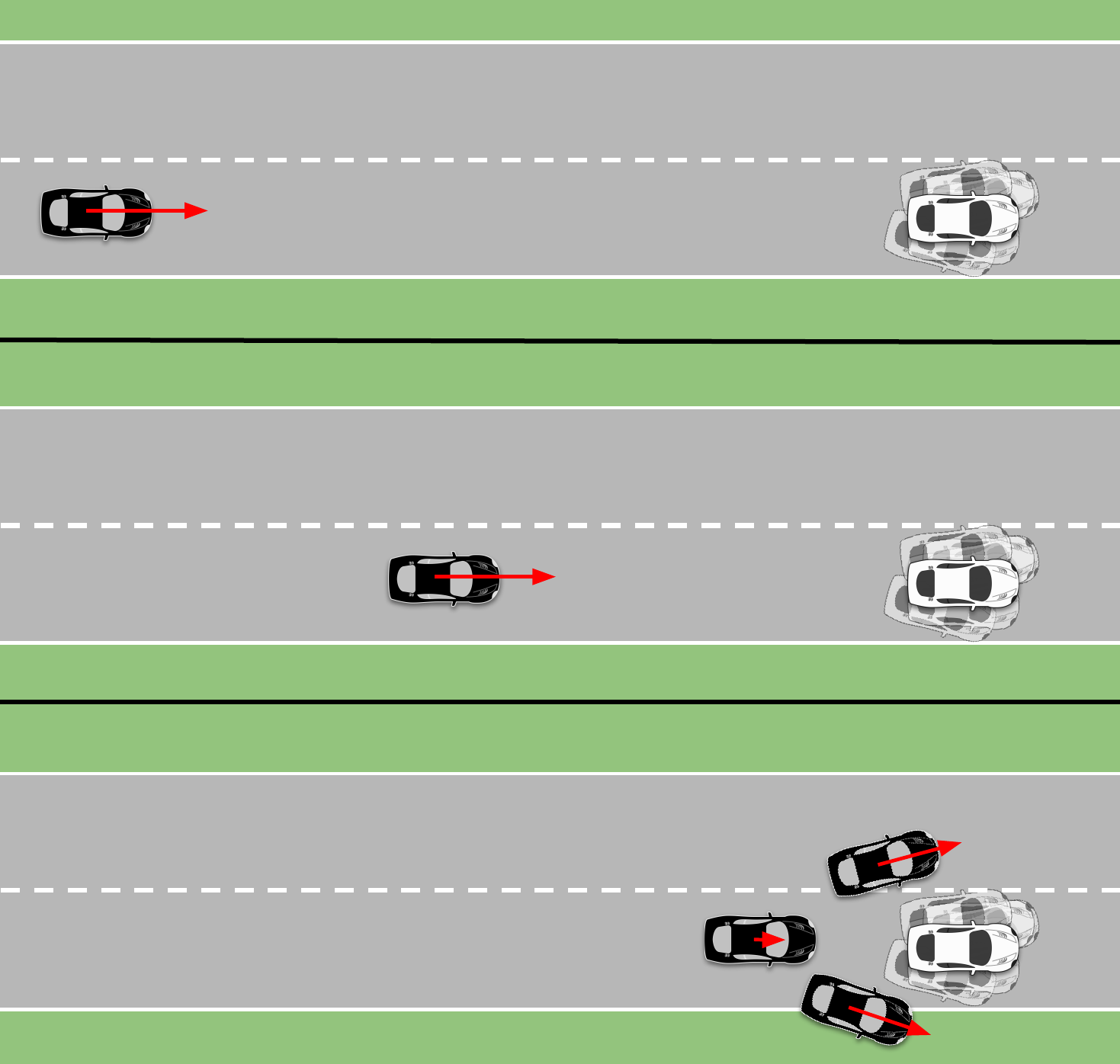}
    \caption{Stationary scenario.}
    \label{fig:scenarios-stationary}
\end{subfigure}
\hfill
\begin{subfigure}{0.32\textwidth}
    \includegraphics[width=\textwidth]{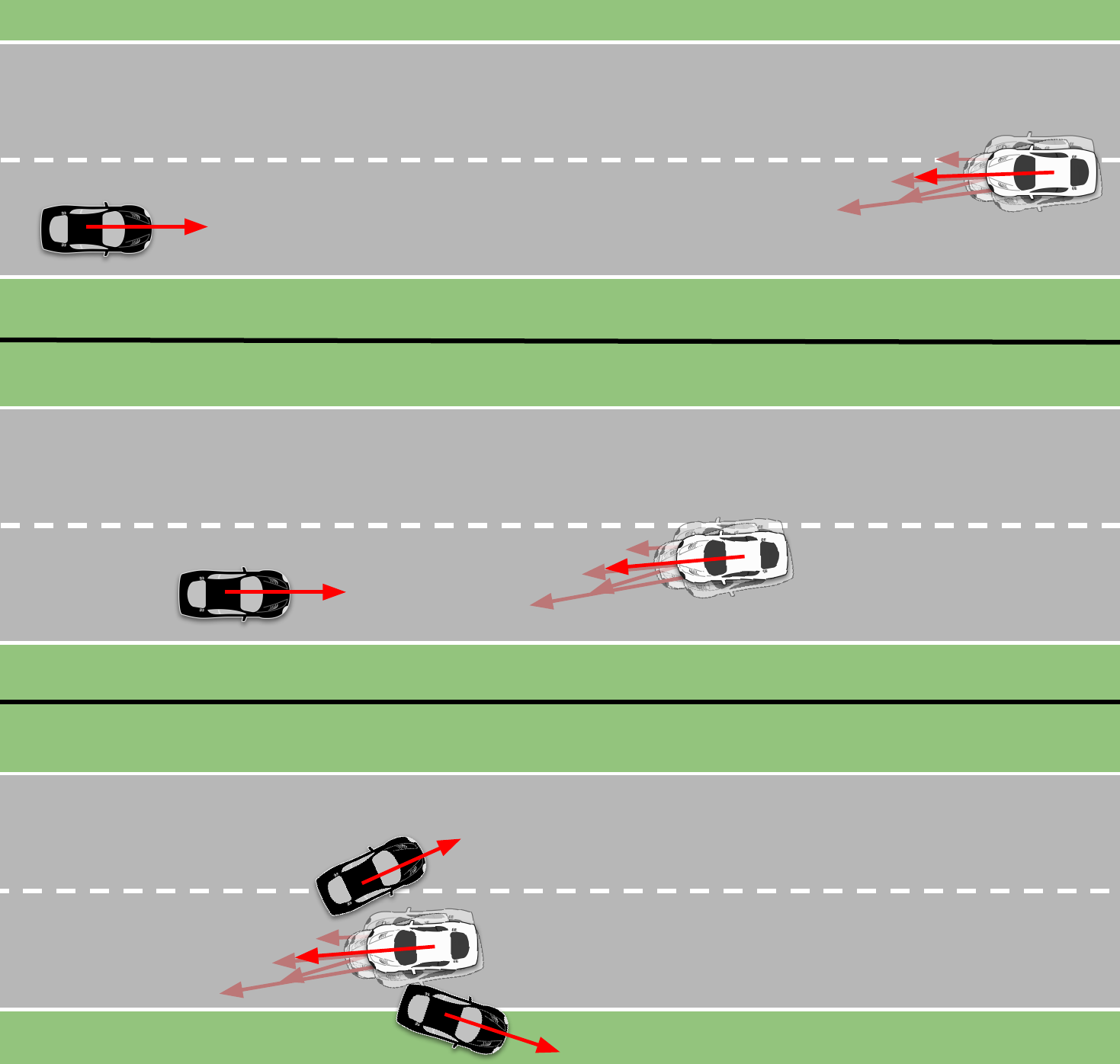}
    \caption{Frontal scenario.}
    \label{fig:scenario-frontal}
\end{subfigure}
\hfill
\begin{subfigure}{0.32\textwidth}
    \includegraphics[width=\textwidth]{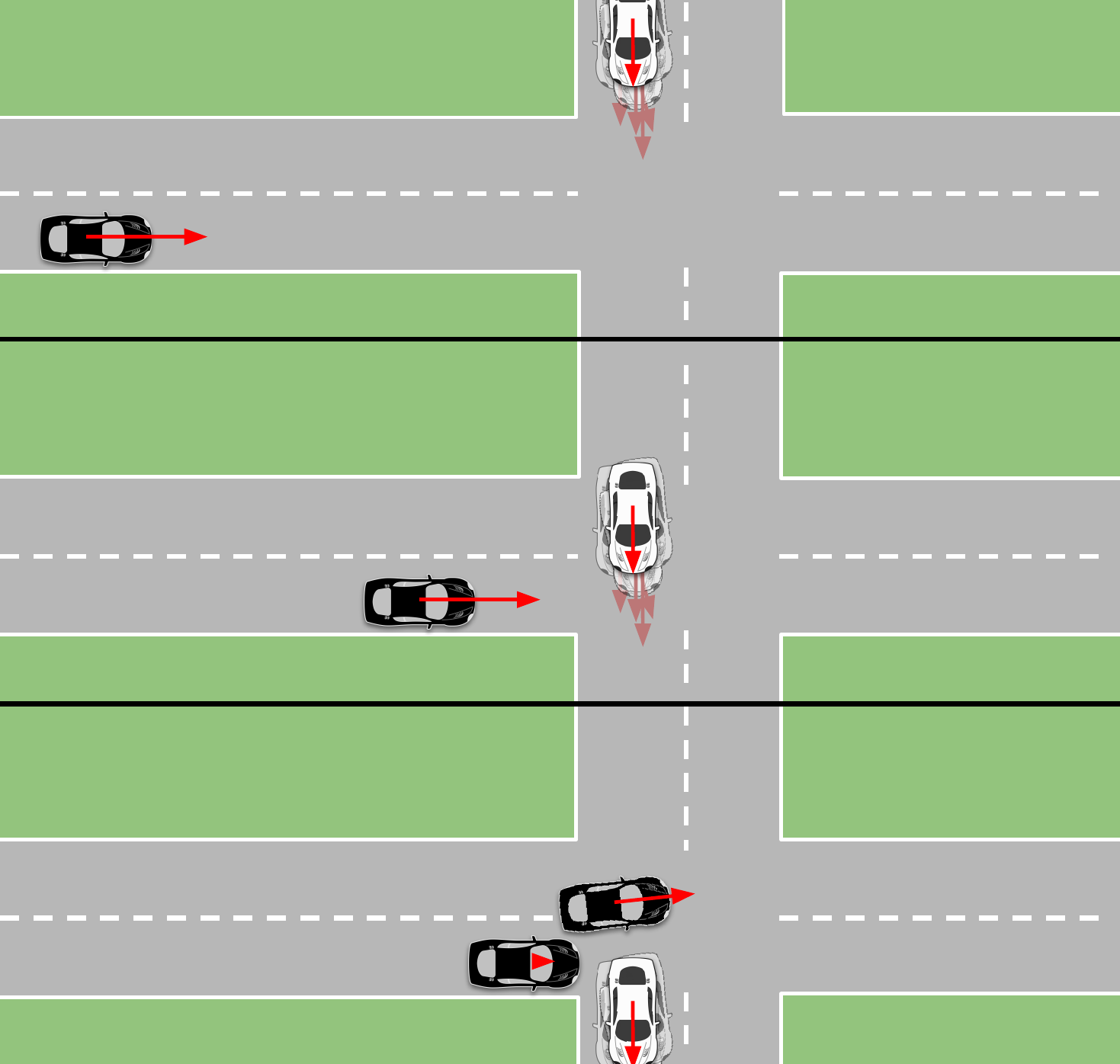}
    \caption{Side scenario.}
    \label{fig:scenarios-side}
\end{subfigure}
        
\caption{Different scenario types used in the NeuroNCAP evaluation protocol. The planner is allowed perceptual input $t_{pre}$ seconds before the test starts in order to build temporal context. Once the test starts, at $t=t_{start}$, there are multiple actions that can lead to a successfully completed scenario e.g., harsh breaking or a steering maneuver. To increase the robustness of the test and allow for multiple runs, we introduce small random perturbations to the target actor. Note that this is an illustration and not the actual output of our renderer.}
\label{fig:scenarios}
\end{figure}

%\red{talk about how we designed and generated our scenarios. Random jittering, random actors, etc. In particular, we can talk about our three types of scenarios:}

%\red{I am missing an explanation of the actual scoring. How do I get my terminal velocity and consequently my NCAP stars. How many variants are there of each test, how do they get progressively harder etc.}

\parsection{Stationary}
This is a relatively simple type of scenario where a stationary target actor has been placed in the ego-lane. The target vehicle can be placed with an arbitrary rotation, but will remain stationary throughout the scenario. This means that the ego-vehicle can either commit to a harsh break or a steering maneuver to avoid a crash. See \cref{fig:scenarios-stationary} for an illustration.

\parsection{Frontal}
The frontal scenarios comprise a target actor that is driving in the opposite direction and has drifted over into the ego-lane on a collision path with the ego-vehicle. Thus, the ego-vehicle cannot avoid crashing by breaking, only reducing the impact speed. To completely avoid a collision, the ego-vehicle must instead perform a steering maneuver. See \cref{fig:scenario-frontal} for an illustration. 

\parsection{Side} 
The side-collision scenarios feature a target actor crossing our lane from a perpendicular direction. If the current velocity of the ego-vehicle is maintained, there will be a side-collision. The ego-vehicle can avoid collision either by braking for the oncoming target actor, or by conducting a slight steering maneuver while speeding past the target actor. See \cref{fig:scenarios-side} for an illustration. 

\parsection{NeuroNCAP score} For each scenario, a score is computed. A full score is achieved only by completely avoiding collision. Partial scores are awarded by successfully reducing the impact velocity. In spirit of the 5-star Euro NCAP rating system\cite{euroNCAP2023collision} we compute the NeuroNCAP score (NNS) as
\begin{equation}
\label{eq:ncap-score}
    \text{NNS} = 
    \begin{cases}
        5.0                                    & \text{if no collision} \\
        4.0 \cdot \text{max}(0, 1 - v_i / v_r) & \text{otherwise}
    \end{cases}\enspace,
\end{equation}
where $v_i$ is the impact speed as the magnitude of relative velocity between ego-vehicle and colliding actor, and $v_r$ is the reference impact speed that would occur if no action is performed. In other words, the score corresponds to a 5-star rating if collision is entirely avoided, and otherwise the rating is linearly decreased from four to zero stars at (or exceeding) the reference impact speed.
\section{Experiments}
First, we start by outlining the details of our experiments in \cref{sec:experiment}. Next, we show the quantitative results from our NeuroNCAP evaluation in \cref{sec:neuroncap-results} and some qualitative examples in \cref{sec:qualitative-results}. Last, we show results from our real-to-sim study in \cref{sec:real-to-sim}, building more confidence in the results from the NeuroNCAP evaluation.

\subsection{Experimental Setting}
\label{sec:experiment}
\parsection{Dataset} While there are many datasets targeting autonomous driving~\cite{alibeigi2023zenseact, mei2022waymo, wilson2023argoverse, xiao2021pandaset, geiger2012we}, nuScenes~\cite{caesar2020nuscenes} has received the most widespread adaptation for end-to-end planning. It features urban environments with highly interactive scenarios, making it suitable for our safety-critical scenario generation. Thanks to its widespread adaptation it also allows us to use official implementations and network weights of the models we evaluate. NuScenes is divided into 1000 sequences, out of which 150 are reserved for validation. From these 150, we choose 14 diverse sequences -- deemed to be suitable based on the behavior of agents present in the scene -- to serve as the basis for our safety-critical scenarios.

\parsection{Scenarios} Each scenario is designed by hand, considering which actors are suitable for the given sequence, the most reasonable collision trajectories, as well as defining allowed ranges for the different kinds of randomization. During evaluation we run each scenario 100 times (with fixed random seed) and average the results. Not all sequences can be used for all types of scenarios, as for instance we cannot simulate a realistic side collision on a single straight road. We therefore select suitable sequences for each scenario type. For more details, and qualitative examples of each scenario, we refer to the supplementary material.

\parsection{Neural renderer} As our renderer, we opt to use NeuRAD~\cite{tonderski2023neurad}, a SotA neural renderer developed specifically for autonomous driving and verified to work well with nuScenes. As we wish to maximize the reconstruction quality, we use the larger configuration (NeuRAD-L), and train for 100k steps with default hyperparameters. As pose information in nuScenes is limited to the bird's eye view plane, we employ pose optimization to recover the missing information. Finally, we adopt actor flipping along the symmetry axis \cite{yang2023unisim} to enable realistic rendering of actors from all viewpoints.

\parsection{AD models}
We evaluate two current SotA end-to-end driving models, namely UniAD~\cite{hu2023planning} and VAD~\cite{jiang2023vad}, according to our proposed evaluation protocol. In both cases, we make use of the pre-trained weights made available by the authors, trained on the same dataset, without any alterations to the configuration of said models. Both of these models consume 360° camera input, along with can-bus signals and a high-level command: \textit{right}, \textit{left}, or \textit{straight}, and output a sequence of future waypoints up to 3 seconds into the future. While this is shorter than the initial time-to-collision (TTC) in our scenarios, it is not an issue as the evasive maneuver can, and should, begin before the final waypoint intersects the current actor position. Additionally, our scenarios are designed to be quite lenient, so that a plan at TTC < 3s can still successfully avoid collision.  

One major difference between these two models is that UniAD applies a collision-avoidance optimization post-processing step to their predicted trajectory. The optimization is performed using a classical solver with a cost-function based on predicted occupancy and the non-optimized output trajectory. This optimization was shown to drastically decrease the collision-rate when evaluated in open loop, and we can now study it in the more interesting closed-loop setting. To enable more directly comparable analysis, we implement the same collision avoidance optimization for VAD. However, as VAD does not directly predict future occupancy, we rasterize their predicted future objects and use this as the future occupancy. Note that this approach possibly overestimates occupancy, as all future modes are treated as equally likely.

For comparison we implement a naïve baseline method based on the perception outputs of UniAD/VAD. The planning logic is simply a constant velocity model unless we observe an object in a corridor in front of the ego-vehicle, in which case we perform a braking maneuver. The corridor is defined as $\pm 2$ meters in the lateral direction and ranging from 0 to $2v_{ego}$ meters in the longitudinal direction, \ie we brake if we have TTC < 2s with an object in front of us.

\subsection{NeuroNCAP Results}
\label{sec:neuroncap-results}
We evaluate VAD~\cite{jiang2023vad} and UniAD~\cite{hu2023planning}, as well as the naïve baseline, on our safety-critical scenarios. We also evaluate both methods with and without perception-based trajectory post-processing. We report the NeuroNCAP score \eqref{eq:ncap-score} and collision rate per scenario type in \cref{tab:neuro-ncap}. Note that the collision rate is not averaged over time, but is defined as the ratio of scenarios that passed without any collisions.

Surprisingly, we find that the plan predicted directly by the network, \ie without post-processing, is extremely unsafe and crashes most of the time, even in the simple stationary scenarios. For reference, the naïve baseline achieves an almost perfect score in the stationary setting, showing both that the perception of these models is not at fault, and that very simple logic can avoid collision. Trajectory post-processing further confirms this, reducing the collision rate dramatically in the stationary setting. Side and frontal scenarios are more difficult to handle with this rule-based logic, and the baseline crashes 100\% of the time, albeit with a lower impact speed (thus scoring higher). Surprisingly, the end-to-end methods again show almost no reaction to the impending collision, with 98-99\% collision rate in frontal scenarios. Trajectory post-processing improves safety somewhat, but is not nearly as effective as in the stationary setting. 

We believe that these results highlight a drastic flaw in the design or training of current end-to-end autonomous driving systems. Reducing the contradictions between the predicted plan and the auxiliary outputs is a promising area of improvement for future end-to-end planners. Notably, VAD actually attempts to address this by using multiple loss terms that directly encourage the model to output a plan that is consistent with its perception and prediction outputs. However, as our experiments show, this alignment step does not generalize well, at least not to this type of safety-critical scenario. 

% \begin{table}[t]
%     \centering
%     \setlength{\tabcolsep}{3pt}
%     \begin{tabular}{l|c|cccc|cccc}
%                             &            & \multicolumn{4}{c|}{NeuroNCAP Score $\uparrow$} & \multicolumn{4}{c}{Collision rate (\%) $\downarrow$}   \\

%             Model          & Post-proc. & Avg. & Stat. & Frontal & Side  & Avg. & Stat. & Frontal  & Side   \\ \hline
%             Base-U         & -          & 2.91 & 4.90 & 2.52 & 1.31 & 64.8 & 3.0 & 98.7 & 92.7 \\
%             Base-V         & -          & 2.89 & 4.92 & 2.44 & 1.31 & 64.5 & 2.1 & 98.7& 92.7 \\ \hline
%             UniAD          & x          & 0.97 & 1.47 & 0.06 & 1.39 & 82.4 & 74.6 & 99.3 & 73.3 \\
%             VAD$^\dagger$  & x          & 0.85 & 0.99 & 0.12 & 1.45 & 86.2 & 87.3 & 98.0 & 73.3 \\ \hline
%             UniAD$^\dagger$& \checkmark & 2.02 & 3.82 & 0.81 & 1.44 & 63.5 & 27.7 & 88.0 & 74.7 \\
%             VAD            & \checkmark & 2.73 & 4.24 & 1.43 & 2.51 & 50.7 & 18.3 & 73.3 & 60.5 
                        
%     \end{tabular}
%     \caption{NeuroNCAP evaluation results. End-to-end planners fail in novel, critical scenarios. Trajectory post-processing, as proposed in UniAD, helps significantly. The naïve baseline uses the perception of either UniAD (U) or VAD (V) to determine braking. $^{\dagger}$Corresponds to the model's original setting.}
%     \label{tab:neuro-ncap}
% \end{table}

\begin{table}[t]
    \centering
    \setlength{\tabcolsep}{3pt}
    \begin{tabular}{l|c|cccc|cccc}
                            &            & \multicolumn{4}{c|}{NeuroNCAP Score $\uparrow$} & \multicolumn{4}{c}{Collision rate (\%) $\downarrow$}   \\

Model          & Post-proc. & Avg. & Stat. & Frontal & Side  & Avg. & Stat. & Frontal  & Side   \\ \hline
Base-U         & -          & 2.65 & 4.72 & 1.80 & 1.43 & 69.90 & 9.60 & 100.00 & 100.00 \\
Base-V         & -          & 2.67 & 4.82 & 1.85 & 1.32 & 68.70 & 6.00 & 100.00 & 100.00 \\ \hline
UniAD          & x          & 0.73 & 0.84 & 0.10 & 1.26 & 88.60 & 87.80 & 98.40 & 79.60 \\
VAD$^\dagger$  & x          & 0.66 & 0.47 & 0.04 & 1.45 & 92.50 & 96.20 & 99.60 & 81.60 \\ \hline
UniAD$^\dagger$& \checkmark & 1.84 & 3.54 & 0.66 & 1.33 & 68.70 & 34.80 & 92.40 & 78.80 \\
VAD            & \checkmark & 2.75 & 3.77 & 1.44 & 3.05 & 50.70 & 28.70 & 73.60 & 49.80 \\
    \end{tabular}
    \caption{NeuroNCAP evaluation results. End-to-end planners fail in novel, critical scenarios. Trajectory post-processing, as proposed in UniAD, helps significantly. The naïve baseline uses the perception of either UniAD (U) or VAD (V) to determine braking. $^{\dagger}$Corresponds to the model's original setting.}
    \label{tab:neuro-ncap}
\end{table}

\subsection{Qualitative Results}
\label{sec:qualitative-results}

\begin{figure*}[t!]
    \centering
    \begin{subfigure}[b]{0.32\textwidth}
        \centering
        \includegraphics[width=\textwidth]{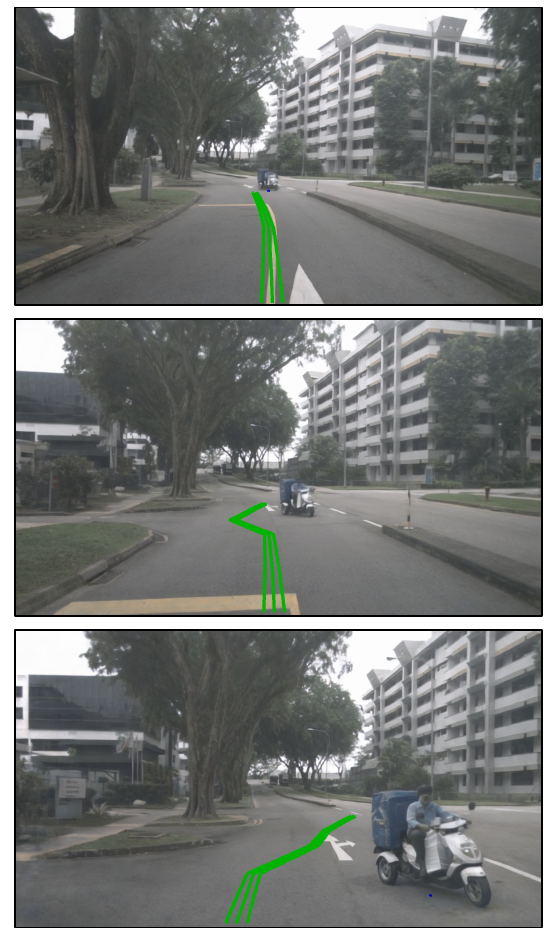}
        \caption{stationary (UniAD)}
        \label{fig:qualitative-a}
    \end{subfigure}%
    ~ 
    \begin{subfigure}[b]{0.32\textwidth}
        \centering
        \includegraphics[width=\textwidth]{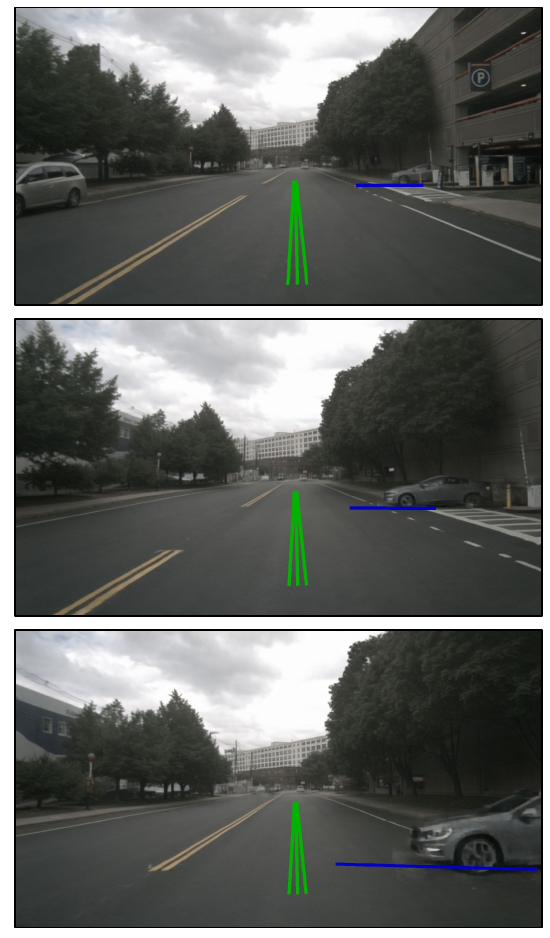}
        \caption{side (VAD)}
        \label{fig:qualitative-b}
    \end{subfigure}% 
    ~ 
    \begin{subfigure}[b]{0.32\textwidth}
        \centering
        \includegraphics[width=\textwidth]{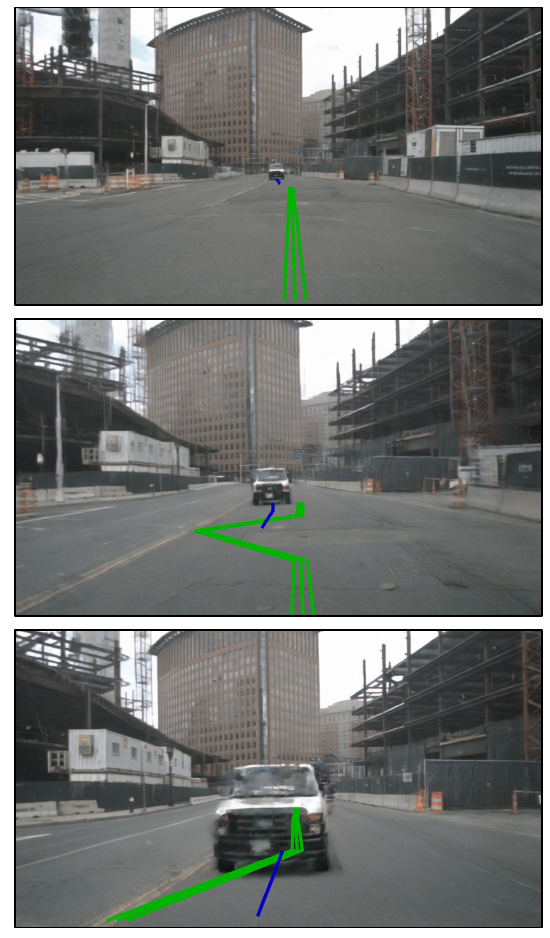}
        \caption{frontal (UniAD)}
        \label{fig:qualitative-c}
    \end{subfigure}%
    
    \caption{Qualitative examples of three NeuroNCAP scenarios, with projected planning output (green, before controller) and the actual designed future trajectory of the target actor (blue). In some cases the planner reacts successfully (a), does not react at all (b), or attempts to avoid collision but fails (c). Our simulator can accurately render complex actors (a), but sometimes exhibits unrealistic artifacts for very close objects (b) and (c).}
    \label{fig:qualitative}
\end{figure*}

We augment the quantitative analysis with rendered front-camera images from each scenario type in \cref{fig:qualitative}, with overlaid projections of the planned trajectories. \cref{fig:qualitative-a} depicts a successful avoidance maneuver, while also highlighting our ability to render complex entities such as a motorcyclist. However, without post-processing, the planners are prone to seemingly ignoring the safety-critical event, as seen in \cref{fig:qualitative-b}.

\begin{figure*}[t!]
    \centering
    \begin{subfigure}[b]{0.32\textwidth}
        \centering
        \includegraphics[width=\textwidth]{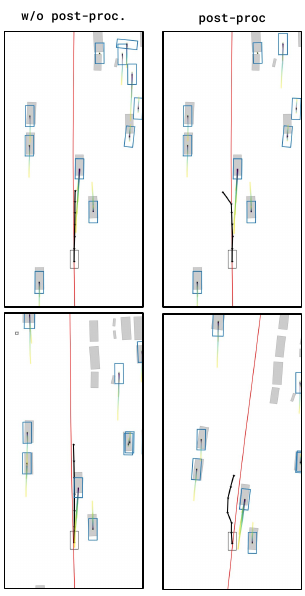}
        \caption{Successful avoidance}
        \label{fig:post-proc-a}
    \end{subfigure}%
    ~ 
    \begin{subfigure}[b]{0.32\textwidth}
        \centering
        \includegraphics[width=\textwidth]{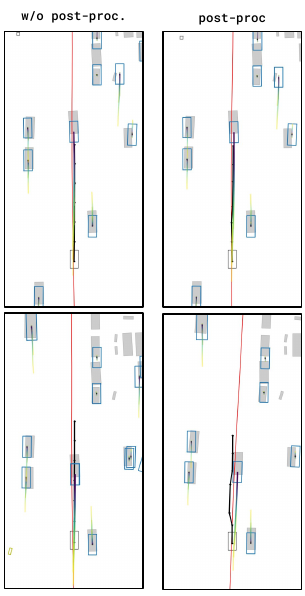}
        \caption{Insufficient margin}
        \label{fig:post-proc-b}
    \end{subfigure}%
    ~ 
    \begin{subfigure}[b]{0.32\textwidth}
        \centering
        \includegraphics[width=\textwidth]{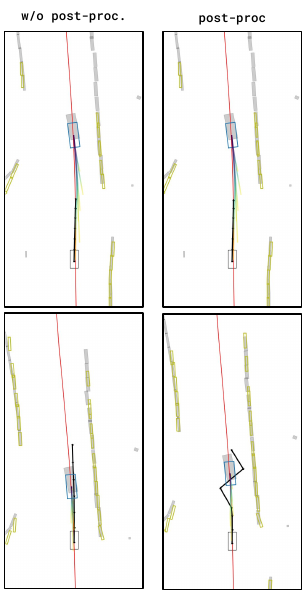}
        \caption{Catastrophic failure}
        \label{fig:post-proc-c}
    \end{subfigure}% 
    \caption{UniAD perception and planning output for three different scenarios, with (right) and without (left) trajectory post-processing. Highlighting unsafe planning despite strong perception, as well as strengths and weaknesses of post-processing. The plot features ground truth objects (grey) and predicted objects (class dependent color), and their predicted future trajectories. Moreover, we show the ego-vehicle (black), its planned trajectory (black) and the reference trajectory it is steering towards (red).}
    \label{fig:post-proc}
\end{figure*}

To further examine this issue, \cref{fig:post-proc} presents the perception and planning outputs across various scenarios, demonstrating that UniAD, without post-processing, frequently plans hazardous trajectories despite robust perception capabilities (left). This indicates that the high collision rate is not due to a real-to-sim gap induced by our renderer, but rather due to the planner not handling the covariate shift between the training data and our safety-critical scenarios. When post-processing is enabled (right), trajectories are adjusted to avoid collisions according to the model's internal perception and prediction outputs. This can result in a few different outcomes. In some cases, such as in \cref{fig:post-proc-a}, this is able to completely prevent a collision by slowing down and/or steering around the target. However, the optimization does not adequately consider the extent of the ego-vehicle, sometimes causing the adjustments to be too small, resulting in glancing collisions as can be seen in \cref{fig:post-proc-b}. Finally, due to not considering the trajectory holistically, and only optimizing each waypoint individually, the result is sometimes catastrophic, as seen in \cref{fig:post-proc-c} and \cref{fig:qualitative-c}. Here, the waypoints are repelled from the object in the opposite direction, actually causing the vehicle to steer into the target actor and accelerate right before impact.

\subsection{Simulation Gap Study}
\label{sec:real-to-sim}
One important -- if not the most important -- concern of doing testing in simulation is to what degree the results transfer to the real world. Therefore, we extensively measure the real-to-sim gap across three crucial facets of the driving system -- perception, prediction, and planning -- in both open and closed-loop settings.

\parsection{Replay real-to-sim (open-loop)}
The authors of NeuRAD~\cite{tonderski2023neurad} show that the perception gap between rendered and real images is very small, at least in terms of 3D object detection and relative depth estimation. However, this study was only partially performed on nuScenes, and crucially, did not consider planning metrics, which is what is most relevant in this work. Therefore, we analyze the open-loop real-to-sim gap of VAD and UniAD on the full sequences that our scenarios are based on. In \cref{tab:open-loop}, we report the standard planning metrics Average Displacement Error (ADE) and Collision Rate (CR) metrics, computed over 1, 2, and 3 seconds into the future. We also present the perception real-to-sim gap in terms of NDS, utilizing the auxiliary detection outputs from each model. Our analysis is limited to VAD's maximum range of $\pm30$m and excludes pedestrians as they are not utilized in any of our scenarios (and explicitly not modeled by our renderer).

The displacement error is very similar between original and rendered images, indicating a minimal real-to-sim gap in the open-loop setting. In terms of collision rate, both VAD and UniAD show similar or lower collision rates on rendered images. In terms of detection performance, UniAD shows practically identical performance on real and simulated data, whereas VAD has a gap of roughly 3 points in terms of NDS. However, we argue that this is within acceptable margins, especially as most evaluated objects are farther away and thus less visible than our inserted target actors.

\begin{table}[t]
    \centering
    \setlength{\tabcolsep}{5pt}
    \begin{tabular}{l|l|ccc|ccc|c}
              &           & \multicolumn{3}{c|}{ADE @ T (m) $\downarrow$} & \multicolumn{3}{c|}{CR @ T (\%) $\downarrow$} &  Detection $\uparrow$                             \\
        Model & Modality  & 1.00s & 2.0s & 3.0s & 1.00s & 2.0s & 3.0s & \hspace{2mm}NDS$^*$\\ \hline
        %UniAD & (reported) & 0.42 & 0.63 & 0.91 &  0.07 & 0.10 & 0.22 \\ % do not use as we have our own evaluation
        UniAD & original  & 0.44 & 0.75 & 1.16 & 0.00 & 0.12 & 0.21 & 0.490 \\
        UniAD & simulated & 0.47 & 0.80 & 1.24 & 0.00 & 0.12 & 0.24 & 0.489\\ \hline
        %VAD & (reported) & 0.41 & 0.70 & 1.005 & 0.07 & 0.17 & 0.41 \\ % do not use as we have our own evaluation
        VAD   & original  & 0.43 & 0.71 & 1.01 & 0.00 & 0.08 & 0.11 & 0.449\\
        VAD   & simulated & 0.44 & 0.76 & 1.16 & 0.00 & 0.00 & 0.08 & 0.413\\
            \end{tabular}
    \caption{Real-to-sim evaluation in open loop on our 14 nuScenes sequences, using typical planning and detection metrics. $^*$Computed over VAD's range for fair comparison between models.}
    \label{tab:open-loop}
    \vspace{-3mm}
\end{table}

\parsection{Scenario real-to-sim (closed-loop)}
The previous study was performed on the original sequences, where ground truth is readily available for perception, prediction, and planning. However, we also study the behavior of the models in our closed-loop scenarios, where we have edited the actors and can observe the scene from novel views. Particularly, we aim to verify that the model is able to accurately perceive and predict the motion of the inserted actor that is causing the criticality of the scenario. Thus, we evaluate metrics related to the estimation of these actors, considering recall at present (0s) for perception and recall at future times for prediction. In calculating recall, we define a true positive as a detection or future prediction with either a non-zero overlap or a center distance smaller than 2 meter with the target actor. This approach is deliberately lenient to distinguish inherent perception uncertainty in 3D estimation from failures induced by rendering artifacts. We compute and average these metrics for all our scenarios and present them on a per-range-to-actor basis in \cref{tab:target-actor-metrics}.

\begin{table}[]
    \centering
    \resizebox{\linewidth}{!}{%
        \begin{tabular}{l|l|ccc|ccc|ccc|ccc}
                                        &       & \multicolumn{3}{c|}{Recall @ 0s $\uparrow$} & \multicolumn{3}{c|}{Recall @ 1s$\uparrow$} & \multicolumn{3}{c|}{Recall @ 2s$\uparrow$} & \multicolumn{3}{c}{Recall @ 3s $\uparrow$}                                                                       \\
            Scenario                    & Model & 5-15m                                       & 15-25m                                     & 25-35m                                     & 5-15m                                      & 15-25m & 25-35m & 5-15m & 15-25m & 25-35m & 5-15m & 15-25m & 25-35m \\ \hline
\multirow{2}{*}{Stationary}&UniAD&0.97&0.98&0.89&0.94&0.95&0.84&0.94&0.93&0.76&0.94&0.90&0.69\\
& VAD& 1.00& 0.96& 0.70& 0.97& 0.87& 0.64& 0.93& 0.82& 0.60& 0.91& 0.80& 0.57\\\hline
\multirow{2}{*}{Frontal}&UniAD&0.83&0.97&0.90&0.82&0.96&0.90&0.80&0.93&0.83&0.77&0.91&0.73\\
& VAD& 0.96& 0.97& 0.70& 0.91& 0.93& 0.69& 0.83& 0.86& 0.63& 0.65& 0.69& 0.54\\\hline
\multirow{2}{*}{Side}&UniAD&0.92&0.96&0.65&0.92&0.96&0.65&0.90&0.95&0.62&0.72&0.63&0.56\\
& VAD& 0.90& 0.64& 0.44& 0.87& 0.64& 0.40& 0.82& 0.62& 0.38& 0.56& 0.51& 0.36\\
        \end{tabular}
    }
    \vspace{0.5mm}
    \caption{Target actor recall in different ranges and at different times in the future. Note that the model consistently has a good understanding of the target actors' dynamics as shown by the high recall during future prediction.}
    \label{tab:target-actor-metrics}
\end{table}

The results in \cref{tab:target-actor-metrics} indicate that the model in most cases has a good understanding of the target actor, its current state, and its potential future states. With high recall scores (>80\%) in the most safety-critical ranges (5-25m to the target-actor), the planner should be able to plan a safe collision avoidance maneuver. Note that VAD has a range of $\pm30$ meters in the longitudinal direction and $\pm15$ meters in the lateral direction. This can be seen in the overall decreased recall rate in the 25-35m range, as well as in the 15-25 meter range in the side scenarios. 
\section{Limitations}
We see the following limitations. First, the neural renderer is limited in the scenes and scenarios, \eg, no rain, it is able to accurately render. Moreover, large deviations in ego-vehicle trajectory and very close objects lead to visual artifacts (see \cref{fig:qualitative}). Second, we adopt a simplified vehicle model, which does not model, \eg, delays, friction, or suspension. Further, we do not consider road surface aspects such as bumps, potholes, gravel, etc. Third, we have adopted a single controller for all models, even though they are tightly coupled. Our evaluation protocol allows for submitting AD models that directly output control signals. Fourth, the neural renderer is unable to deal with deformable objects, such as pedestrians. We hope that further advances in neural rendering will lift this restriction and enable a new set of safety-critical scenarios focusing on vulnerable road users. Fifth, the target actor follows a predetermined trajectory, without dynamically reacting to the ego-vehicle.
While this follows the Euro NCAP setting, we believe that future scenarios with multiple actors would require reactive behavior.

\section{Conclusion}

In conclusion, our simulation environment offers a novel approach for evaluating the safety of autonomous driving models, drawing on real-world sensor data and Euro NCAP-inspired safety protocols. Through the NeuroNCAP framework, which includes stationary, frontal, and side collision scenarios, we have exposed significant vulnerabilities in current SotA planners. These findings not only underline the urgent need for advancements in the safety of end-to-end planners but also suggest promising paths for future research. By making our evaluation suite openly available to the wider research community, we aim to catalyze progress towards safer autonomous driving. Looking ahead, we anticipate evolving the suite to tackle a wider range of scenarios, integrating more refined vehicle models, and employing advanced neural rendering techniques, thereby setting new benchmarks for safety evaluation.

% Added acknowledgements for WASP cam-ready
\section*{Acknowledgements}
We thank Georg Hess, Carl Lindström, and Adam Lilja for fruitful discussions. This work was partially supported by the Wallenberg AI, Autonomous Systems and Software Program (WASP) funded by the Knut and Alice Wallenberg Foundation. Computational resources were provided by NAISS at \href{https://www.nsc.liu.se/}{NSC Berzelius}, partially funded by the Swedish Research Council, grant agreement no. 2022-06725.

\bibliographystyle{splncs04}
\bibliography{main}
\end{document}